%% file: paper.tex
\documentclass[11pt]{humain-lab}

\input{macros}

\title{Collaborative Theory of Mind: Learning to Self-Revise by Anticipating User Responses}

\author[1]{Fanzhi Zeng}
\author[2]{Anubrata Das}
\author[1]{Maytal Saar-Tsechansky}
\author[1,\dagger]{Liu Leqi}

\affiliation[1]{The University of Texas at Austin}
\affiliation[2]{Autodesk}

\contribution[\dagger]{Corresponding author}

\input{sections/00_abstract}

\correspondence{Fanzhi Zeng at \email{fanzhi.zeng@utexas.edu}, Liu Leqi at \email{leqiliu@utexas.edu}}
\codeurl{https://github.com/thuzfz1/CToM}

\reportdate{October 2026}

\begin{document}

\maketitle

\input{sections/01_introduction}

\input{sections/02_related_work}

\input{sections/03_problem_setup}

\input{sections/04_method}
\input{sections/05_experiments}
\input{sections/06_discussion}

\section*{Acknowledgement}
This research is supported in part by funding from Coefficient Giving. Computational resources are provided by the Texas Advanced Computing Center (TACC).

\bibliographystyle{assets/plainnat}
\bibliography{citation}

\clearpage
\newpage
\appendix
\input{sections/09_appendix}

\end{document}

%% file: macros.tex
\usepackage{fvextra}              %
\usepackage{upquote}              %
\usepackage{fontawesome5}

\newcommand{\methodname}{\textsc{CToM}}

\newcommand{\step}[1]{\ding{\numexpr171+#1\relax}}   %

\definecolor{satgreen}{HTML}{00A650}
\definecolor{unsatorange}{HTML}{C55A11}
\newcommand{\satisfied}{\textcolor{satgreen}{\faSmile[regular]}\,\textsc{Satisfied}}
\newcommand{\unsatisfied}{\textcolor{unsatorange}{\faFrown[regular]}\,\textsc{Unsatisfied}}

\newcommand{\pmsem}[2][]{\textsubscript{\textcolor{black!55}{$\pm$#2}\phantom{#1}}}

\newcommand{\promptsize}{\fontsize{8}{9.5}\selectfont}  %
\newcounter{aiprompt}
\newtcolorbox{aipromptboxinner}[1]{title={Prompt~\theaiprompt: #1}}
\newenvironment{aipromptbox}[1][]%
  {\refstepcounter{aiprompt}\begin{aipromptboxinner}{#1}}%
  {\end{aipromptboxinner}}
\newcommand{\promptref}[1]{Prompt~\ref{#1}}
\newcommand{\promptsref}[2]{Prompts~\ref{#1}--\ref{#2}}

\makeatletter
\newcommand{\analysisfigurecaption}{\def\@captype{figure}\caption}
\makeatother

%% file: sections/00_abstract.tex
\abstract{
Large language models (LLMs) demonstrate strong individual task performance, yet still
struggle to collaborate with users over extended, multi-turn interactions, for example by
attempting solutions before understanding the user's needs or by failing to revise mistaken
assumptions after receiving feedback. We propose Collaborative Theory of Mind
(\methodname{}), a framework that addresses these challenges by explicitly training LLMs to
anticipate user responses and to use those predictions to improve their own behavior.
\methodname{} drafts a response, predicts how the user will respond, and uses this prediction
to guide internal revision before presenting a response. During training, the user's reply to
each draft serves as the target for predicting the next turn, and we train the assistant to
minimize the negative log-likelihood of that reply. We combine this prediction objective with
group-relative policy optimization (GRPO) to optimize both task performance and interaction
efficiency.
Across mathematical reasoning, coding, and writing tasks, \methodname{} achieves higher
collaborative task performance than all baselines; for example, on a collaborative mathematical
reasoning task, it improves final accuracy from 74.1\% to 79.6\% compared with naive GRPO.
Ablations show that predicting user responses also makes collaboration more efficient: under the
same reward, adding prediction supervision reduces the number of user replies by 8\% on coding and
17\% on writing while improving task performance.
Behavioral analyses of interaction trajectories show that user-response predictions become more accurate after training, and that more accurate predictions are associated with larger improvements in collaborative task performance through self-revision.
}

%% file: sections/01_introduction.tex
\section{Introduction}
\label{sec:intro}

\begin{figure}[t]
  \centering
  \includegraphics[width=.9\linewidth]{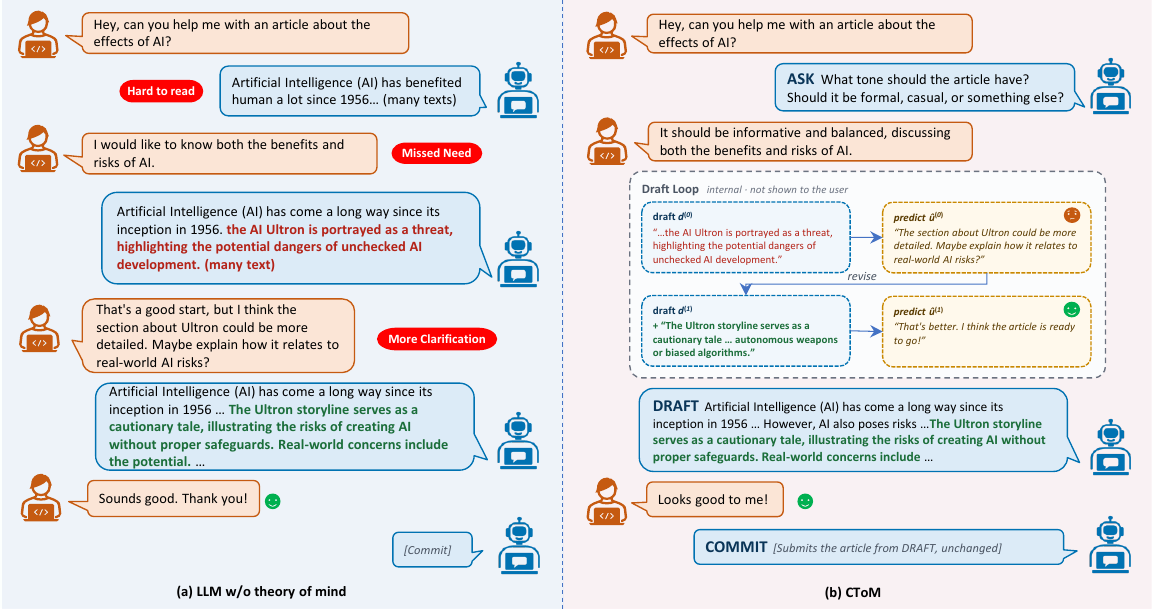}
  \caption{A real example of an LLM without theory of mind and \methodname{} on the
  same request. \textbf{(a)}~Without a model of the user, the LLM answers at once with a long
  draft that misses the user's need, and the user must correct it twice. \textbf{(b)}~\methodname{}
  first asks about the tone, then drafts internally: its predicted reply to the first draft flags
  the weak Ultron section, so it revises before showing anything, and the user accepts the first
  visible draft. The user of (a) reads three long drafts and writes two corrections, while
  \methodname{} anticipates the problem and saves effort from the user.}
  \label{fig:teaser}
\end{figure}

Today's large language models (LLMs) demonstrate strong capabilities in mathematical reasoning, code generation, and writing, motivating their use as assistants for increasingly complex tasks. In practice, many of these tasks require multiple exchanges in which users and assistants clarify requirements, develop solutions, and respond to feedback. For example, a user drafting a project proposal may refine its scope, intended audience, and priorities while reviewing successive drafts. Strong performance on individual tasks, however, does not reliably translate into effective collaboration. LLMs can attempt solutions before understanding the user's needs, make assumptions that
require repeated correction, and struggle to recover when their initial approach proves
unsuccessful (Figure~\ref{fig:teaser}a)~\citep{laban2025lost,wu2025collabllm}. %

Effective collaboration requires reasoning about the person one is interacting with. 
In human collaboration, people rely on \emph{theory of mind}---understanding what others know,
want, and intend---to anticipate how they will act and react~\citep{premack1978does}.
This ability makes a measurable difference: groups whose members score
higher on a theory-of-mind test perform better on a broad category of tasks, even when
they communicate only through text \citep{engel2014reading}. 
A collaborator preparing a
draft, for instance, can consider what the recipient may find unclear with
their goals and address those issues before sharing it. 
This perspective motivates a concrete training principle for AI assistants: learning to anticipate how a user would respond can help an assistant identify and address shortcomings in its own response before sharing it with the user. 

Inspired by this, we propose \textbf{Collaborative Theory of Mind (\methodname{})}, which
trains an assistant to predict the user's next reply to its own response and to act on that
prediction (Figure~\ref{fig:teaser}b). As shown in Figure~\ref{fig:overview}, the assistant
drafts a response, predicts
the user's feedback on it, and revises the draft in an internal loop until it expects the user
to be satisfied or reaches its revision budget; only then does it show the response. During
training, the user's real reply to each draft supervises the prediction for that draft. We train the prediction with a supervised loss and the task behavior with
group-relative policy optimization (GRPO)~\citep{shao2024deepseekmath}. \methodname{} differs from prior work in two ways.
First, prior work on theory of mind in LLMs primarily evaluates this ability through explicit reasoning about users' beliefs and profiles~\citep{kim2023fantom,ruan2026beyond}; we train assistants to develop this ability by anticipating user responses, with the goal of improving collaboration. 
Second, prior methods use simulated future conversations to reward collaborative outcomes~\citep{wu2025collabllm} or revise drafts using the model's own critiques~\citep{madaan2023selfrefine}. In \methodname{}, we explicitly train the assistant to predict how the user would respond and revise from that perspective, encouraging the assistant to better understand the user while learning to accomplish the task.

We evaluate \methodname{} on collaborative mathematical reasoning, coding, and writing tasks,
training a Qwen3-8B assistant~\citep{yang2025qwen3} with a Qwen3-32B user simulator.
\methodname{} improves the task score over the strongest baseline by 7.5\%, 4.4\%, and 6.5\% on
math, coding, and writing, respectively. The analyses yield three further findings. First,
predicting user responses improves both task performance and efficiency. Ablating prediction
supervision and the effort penalty separately, we find that prediction supervision raises the task
score in every comparison, and that \methodname{}, which uses both, requires the fewest user input effort among these variants.
Second, predictions of user replies become more accurate after training, and these improved predictions help the assistant revise responses more effectively, both within the internal revision loop and across conversation turns.
Third, forcing models to ask more
clarification questions does not automatically improve performance; encouraging the assistant to ask more questions
lowers its task score in every domain.

Our contributions are as follows:
\begin{itemize}[leftmargin=3.5mm,topsep=2pt,itemsep=1pt,parsep=0pt]
  \item We propose \methodname{}, a training framework that makes predicting the user's next
  reply an explicit learning objective and uses these predictions in an internal revision
  loop, so that the assistant corrects its responses before the user sees them.
  \item We evaluate \methodname{} on mathematical reasoning, coding, and writing, where it outperforms all baselines in task performance. Through ablations, we show that predicting user responses also reduces the effort users spend, measured by the number and length of their replies.
  \item We show that our training objective teaches the assistant to anticipate user replies and use these predictions to revise responses more effectively. 
The objective also yields interesting interaction strategies. For example, the learned policy outperforms one explicitly encouraged to ask more clarifying questions, despite having no direct training signal for that. 
\end{itemize}

%% file: sections/02_related_work.tex
\section{Related Work}
\label{sec:related}

\paragraph{Interactive Evaluation}

Human interaction is seldom one-shot. 
Underspecification is a natural aspect of human conversation~\citep{herlihy2024overcoming}.
Naturally, over the course of human-LLM interaction, user intent, goal, and task completion are revealed.
Pre-LLM literature on dialogue systems argues that considering both task outcomes and the interaction is instrumental~\citep{walker1997paradise}.
Moreover, task success does not necessarily translate to improved human-LLM interaction~\citep{lee2023evaluating}.
Recent work in multi-turn evaluation has highlighted that LLM performance degrades when user intents get clearer over multiple turns \citep{laban2025lost}.
Their analysis shows that LLMs tend to make premature assumptions. 
Overreliance on these early assumptions is a key failure mode as the model fails to incorporate information from late interaction in a conversation episode.
These works motivate that human-LLM interaction needs to be evaluated both in terms of interaction quality and final task performance~\citep{imai2026task}.

\paragraph{Collaborative LLMs}

A growing body of research on collaborative language models investigates how effective LLMs are as collaborators and how they can be trained to be better collaborators. 
CoAuthor analyzes detailed traces of human-AI writing sessions and highlights that notions of good collaboration are grounded in a combination of ownership and productivity \citep{lee2022coauthor}.  
\citet{davidson2026collaboration} identify that models with strong individual performance can perform substantially worse when coordinating between multiple agents.~\citet{shao:Collabor61:online} proposes a collaboration layer where AI can serve as a proactive collaborator.
Complementing the system design aspect, CollabLLM focuses on training models to be proactive collaborators. It uses simulated interactions and multiturn-aware rewards to train models to uncover user intent and make suggestions that improve subsequent outcomes \citep{wu2025collabllm}. Similarly \citet{sun2025training} trains a model to be proactive and personalized beyond task support. In contrast to specific rewards for different aspects of collaboration, we propose a general purpose methodology that infers user intent and revises its output proactively.

\paragraph{Theory of Mind}

LLMs have been shown to exhibit \textit{Theory of Mind}, i.e., the models are able to infer users' mental states without explicit verbalization.
Specialized networks or agents can learn to infer agents' characteristics and mental states from behavioral observations ~\citep{rabinowitz2018machine,rescorla2015computational}. On the other hand, Symmetric Machine Theory of Mind studies such modeling in environments where agents can both communicate and act~\citep{sclar2022symmetric}. 
FANToM extends evaluation to conversational contexts with unequal access to information, revealing difficulties in consistently tracking participants'  beliefs~\citep{kim2023fantom}. ToM-SWE maintains a model of user goals, constraints, and preferences to guide agents~\citep{zhou2025tomswe}. \citet{ruan2026beyond} evaluate and train agents to detect and resolve discrepancies between users' beliefs and the task environment. 
Furthermore, \citet{gong2026does} show that ToM can show improved performance in static benchmarks, however, such improvement may not translate to dynamic environments. RL training on models in collaborative tasks may help with implicit theory of mind capabilities~\citep{sclar2022symmetric}, which provides a formative ground for our method on collaborative theory of mind.

%% file: sections/03_problem_setup.tex
\section{Problem Setting}
\label{sec:setup}

\paragraph{Collaborative interaction.}
We study an LLM assistant that completes a user's task through a multi-turn conversation. A task instance consists of an initial request $u_0$ and task information $z$, such as the user's requirements and goals unspecified to the assistant, available to the user but not directly to the assistant. 

Before assistant turn $t$, the user-visible dialogue is $h_t=(u_0,a_0,u_1,\ldots,a_{t-1},u_t)$, with $h_0=(u_0)$, where $a_i$ and $u_i$ denote assistant and user messages. The assistant policy $\pi_\theta$ produces a response $a_t\sim\pi_\theta(\cdot\mid h_t)$, and if the interaction continues, the user replies $u_{t+1}\sim U(\cdot\mid h_t,a_t,z)$, where $U$ denotes the user's response distribution. The conversation ends according to the task's stopping rule or when its interaction budget is exhausted.

\paragraph{Collaboration objective.}
Let $\mathcal D$ denote the distribution of task instances and $\tau$ a complete rollout of the assistant--user interaction. We seek a policy that maximizes expected conversation return:
\begin{equation*}
 \max_\theta\; J(\theta)
 =\mathbb E_{(u_0,z)\sim\mathcal D,\,\tau\sim(\pi_\theta,U)\mid(u_0,z)}
 \bigl[R(\tau,z)\bigr].
 \label{eq:collaboration-objective}
\end{equation*}
The return $R$ is a task-specific measure of collaboration quality. In general, it may combine
task success with any cost that matters to the user, such as effort, time, or personal experience.
In this paper, we instantiate $R$ as a combined reward of task performance and with a penalty
on interaction effort, measured by the number of turns and the length of user-visible messages
(Section~\ref{sec:method}).

%% file: sections/04_method.tex
\section{Method: Collaborative Theory of Mind (\methodname{})}
\label{sec:method}

In a collaboration, much of the cost of a flawed response falls on the user, who must read it,
notice the problem, and spend a turn explaining it. Many such corrections are predictable from
the conversation so far: a misread constraint, an overlooked requirement, a format the user
already asked for. An assistant that anticipates the user's reply to a candidate response can
fix these problems before the user sees them, saving both a correction turn and the effort of
reading a flawed answer. When the assistant lacks information only the user has, however,
predicting the reply means guessing what it does not know, and revising on that guess would make
it overconfident rather than correct; the right move is to ask.

\begin{figure}[t]
  \centering
  \includegraphics[width=.9\linewidth]{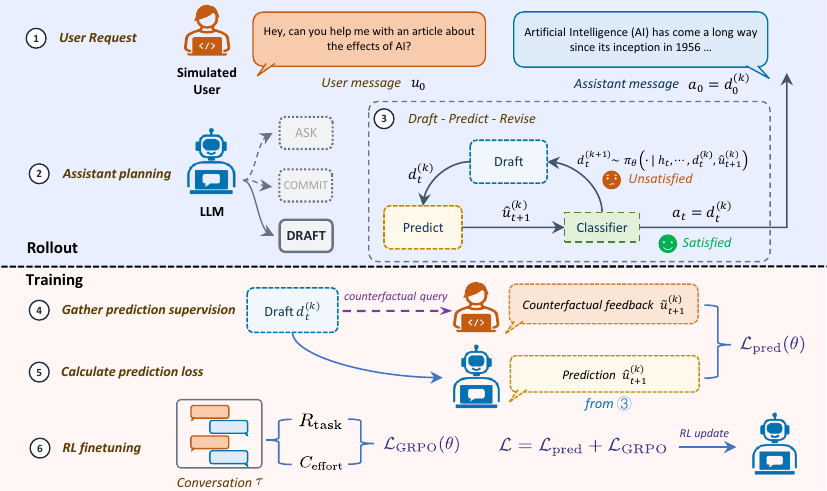}
\caption{Overview of the \methodname{} training pipeline. \emph{Rollout:} given the user's
request $u_0$~\step{1}, the LLM chooses an action~\step{2}; when it drafts, it predicts the
user's reply $\hat{u}_{t+1}^{(k)}$ to each draft $d_t^{(k)}$ and revises until its prediction is
classified as satisfied, then sends $a_t = d_t^{(k)}$ to the user~\step{3}. The user replies,
and this repeats until the conversation $\tau$ ends. \emph{Training:} each discarded draft is
sent to the simulated user in a counterfactual query~\step{4}; its reply, or the user's actual
reply for the shown draft, is the target $\tilde{u}_{t+1}^{(k)}$ that supervises the prediction
$\hat{u}_{t+1}^{(k)}$ through $\mathcal{L}_{\mathrm{pred}}$~\step{5}. The conversation $\tau$
defines $\mathcal{L}_{\mathrm{GRPO}}$, and the sum of the two losses updates the LLM~\step{6}.}
  \label{fig:overview}
\end{figure}

\methodname{} builds both behaviors into a single policy (Figure~\ref{fig:overview}) that chooses
among three actions each round (Section~\ref{sec:method:interaction}). \textsc{Ask} sends a
clarifying question. \textsc{Draft} runs a hidden loop that drafts, predicts the user's reply, and
revises until the prediction signals satisfaction, then shows the final draft. \textsc{Commit}
submits the last shown draft. During training, each prediction is supervised by the simulated
user's reply to that draft: the actual reply for the shown draft and a counterfactual reply for
each discarded one (Section~\ref{sec:method:prediction}). The policy is optimized with GRPO on a
combined reward of task success and a penalty on user effort (Sections~\ref{sec:method:cost}--\ref{sec:method:training}).

\subsection{Prediction-Guided Interaction}
\label{sec:method:interaction}

Each round, the assistant first writes a short plan and then decides how to proceed (system prompts in \promptsref{prompt:ctom-math}{prompt:categorical}). We describe this process at two levels, following Figure~\ref{fig:overview}.

\paragraph{Level 1: choosing an action.}
After planning, the policy samples an action (step~2 in Figure~\ref{fig:overview}). On
\textsc{Ask}, the clarifying question is shown directly as $a_t$, and the user's actual reply
follows without any internal prediction. On \textsc{Commit}, the most recently shown draft
becomes the final answer and the conversation ends; no new message is generated, and
\textsc{Commit} is unavailable before a draft has been shown. On \textsc{Draft}, the assistant
runs the loop below.

\paragraph{Level 2: draft, predict, and optionally revise.}
On a \textsc{Draft} round $t$, the assistant works from the dialogue history $h_t$. The index
$k$ counts internal revisions, starting from zero. At each step, the assistant writes a draft
$d_t^{(k)}$ and predicts the user's reply $\hat u_{t+1}^{(k)}$ to it (step~3):
\begin{equation}
 \begin{aligned}
 d_t^{(k)} &\sim \pi_\theta\big(\cdot \mid h_t,\, d_t^{(0)}, \hat u_{t+1}^{(0)}, \ldots,
            d_t^{(k-1)}, \hat u_{t+1}^{(k-1)}\big),\\
 \hat u_{t+1}^{(k)} &\sim \pi_\theta\big(\cdot \mid h_t,\, d_t^{(0)}, \hat u_{t+1}^{(0)}, \ldots,
            d_t^{(k-1)}, \hat u_{t+1}^{(k-1)},\, d_t^{(k)}\big),
 \end{aligned}
 \label{eq:draft-prediction}
\end{equation}
so each new draft sees all earlier drafts and predictions of the round. A fixed, task-specific classifier labels each predicted reply as \satisfied{} or \unsatisfied{}. If $\hat u_{t+1}^{(k)}$ is \unsatisfied{} and $k<K_{\max}$, the assistant writes a revised draft $d_t^{(k+1)}$; otherwise it shows $a_t=d_t^{(k)}$. Here $K_{\max}$ is the revision cap, and internal drafts and predictions are invisible to the user.

\subsection{Learning to Predict User Feedback}
\label{sec:method:prediction}

Each prediction $\hat u_{t+1}^{(k)}$ in Eq.~\eqref{eq:draft-prediction} is a behavioral model of
the user: it anticipates a correction, a request for clarification, or acceptance of a
particular draft $d_t^{(k)}$. To train these predictions, we need to know how the user would
actually reply to each draft. The user, however, replies only to the draft that is shown.
For the shown draft $a_t$, the target is therefore the user's actual reply $u_{t+1}$. For each
earlier draft discarded during the loop, which the user never sees, we ask the simulated user
after the rollout ends how it would have replied:
\begin{equation*}
 \tilde u_{t+1}^{(k)}\sim U\big(\cdot\mid h_t,\, d_t^{(0)},\hat u_{t+1}^{(0)},\ldots,
 d_t^{(k-1)},\hat u_{t+1}^{(k-1)},\, d_t^{(k)},\, z\big).
 \label{eq:counterfactual-target}
\end{equation*}
We call $\tilde u_{t+1}^{(k)}$ the counterfactual reply. 
The query conditions the simulated
user on everything the assistant produced in the round up to $d_t^{(k)}$, the same context as
the prediction it supervises. 
The assistant never sees these replies, %
which serve only as training targets (step~4 in Figure~\ref{fig:overview}).
\textsc{Ask} and \textsc{Commit} rounds make no predictions and receive no prediction targets.

For notational simplicity, we set $\tilde u_{t+1}^{(k)} = u_{t+1}$ for the shown draft, so that
every draft has a target of the same form. We train the assistant to predict these targets by
minimizing their negative log-likelihood under the same context as the prediction itself,
\begin{equation}
 \mathcal L_{\mathrm{pred}}(\theta)
 =-\frac{1}{|\mathcal S|}\sum_{(t,k)\in\mathcal S}\frac{1}{|\tilde u_{t+1}^{(k)}|}
 \log\pi_\theta\big(\tilde u_{t+1}^{(k)}\mid h_t,\,d_t^{(0)},\hat u_{t+1}^{(0)},\ldots,d_t^{(k)}\big),
 \label{eq:prediction-nll}
\end{equation}
where $\mathcal S$ is the set of drafts in the training batch with targets and
$|\tilde u_{t+1}^{(k)}|$ is the number of tokens in the target. The context keeps the
assistant's own earlier predictions, matching what the assistant
saw when it made the prediction.

\subsection{Interaction Cost}
\label{sec:method:cost}

We measure the interaction a conversation costs the user by the number of user turns $n$ and
the total tokens $\ell_u$ and $\ell_a$ of the visible user and assistant messages over the
conversation; internal drafts and predictions are not counted. We normalize the turn count by
the maximum number of turns $N$, and the token totals by the per-turn token budgets $B_u$ and
$B_a$, clipping each token ratio at $1$ so that all three terms lie in $[0,1]$ and are
comparable. We combine them as
\begin{equation}
C_{\text{effort}}(\tau) = \alpha \frac{n}{N}
  + (1-\alpha-\beta) \min\!\Big(1, \frac{\ell_u}{B_u}\Big)
  + \beta \min\!\Big(1, \frac{\ell_a}{B_a}\Big),
\label{eq:effort}
\end{equation}
where $\alpha,\beta\ge0$ with $\alpha+\beta\le1$. %
Appendix~\ref{app:method-settings} lists $\alpha$, $\beta$, and the task-specific maxima.

\subsection{Combined Reward}
\label{sec:method:return}

We combine the task evaluator's conversation-level score $R_{\mathrm{task}}(\tau, z) \in [0,1]$
(defined for each domain in Appendix~\ref{app:method-settings}) with the interaction cost
$C_{\mathrm{effort}}(\tau)$ from Eq.~\eqref{eq:effort} into the conversation-level reward
\begin{equation}
 R(\tau,z)=w_{\mathrm{task}}R_{\mathrm{task}}(\tau,z)-w_{\mathrm{effort}}C_{\mathrm{effort}}(\tau),
 \label{eq:method-return}
\end{equation}
with $w_{\mathrm{task}}=1$ and $w_{\mathrm{effort}}=0.01$. Both terms are computed once for the
complete conversation. Prediction quality and improvements between internal drafts receive no
separate reward.

\subsection{Joint Training and Inference}
\label{sec:method:training}

\paragraph{From reward to policy optimization.}
For each task instance $(u_0,z)\sim\mathcal D$, we sample a group of $G$ conversations
$\tau_1,\dots,\tau_G$ with the current policy and the simulated user, and then collect the
counterfactual replies to their discarded drafts. GRPO~\citep{shao2024deepseekmath} scores each
conversation relative to its group:
\begin{equation}
 \widehat A_i=\frac{R(\tau_i,z)-\bar R}{\sigma_R+\epsilon},
 \label{eq:reward-advantage}
\end{equation}
where $\bar R$ and $\sigma_R$ are the mean and standard deviation of the rewards in the group,
and $\epsilon>0$ avoids division by zero. The advantages are held fixed during the policy
update, and we use them in the standard clipped GRPO objective
$\mathcal L_{\mathrm{GRPO}}(\theta)$ (Appendix~\ref{app:grpo-details}).%

\paragraph{Joint objective and gradient cap.}
At each update, we minimize
\begin{equation}
 \mathcal L(\theta)=\mathcal L_{\mathrm{GRPO}}(\theta)
 +\eta\,\lambda_{\mathrm{pred}}\mathcal L_{\mathrm{pred}}(\theta),
 \label{eq:joint-objective}
\end{equation}
where the GRPO term optimizes task performance and interaction cost and includes a KL penalty (Appendix~\ref{app:grpo-details}); the prediction term learns
from the targets in Eq.~\eqref{eq:prediction-nll}, and $\eta\in[0,1]$ scales the prediction
contribution.
The prediction loss is a dense, token-level signal, whereas GRPO provides a single scalar per
conversation, so without a limit the prediction gradient could dominate the policy update. To
keep policy optimization the primary objective, we choose $\eta$ at each update as the largest
value for which the prediction gradient is at most half the gradient of the GRPO term:
\begin{equation}
 \eta=\min\!\left(1,\;
 \frac{\big\|\nabla_\theta\mathcal L_{\mathrm{GRPO}}\big\|_2}
      {2\,\lambda_{\mathrm{pred}}\,\big\|\nabla_\theta\mathcal L_{\mathrm{pred}}\big\|_2}\right),
 \label{eq:prediction-scale}
\end{equation}
with $\eta=1$ for a zero prediction gradient and $\eta$ held constant when differentiating Eq.~\eqref{eq:joint-objective}.
Setting $\lambda_{\mathrm{pred}}=0$ removes
prediction supervision while keeping the same prompt and revision loop.

\paragraph{Inference.}
At test time, the assistant runs the same draft–predict–revise loop without any training signal: it decides
whether to revise using its own predictions and sees the user's reply only after showing
$a_t$. Counterfactual queries and the prediction loss are used only during training. Hidden
drafts and predictions do not count toward interaction cost because the user never reads them.

%% file: sections/05_experiments.tex
\section{Experiments}
\label{sec:experiments}
We evaluate \methodname{} on mathematical problem solving, code generation, and document
writing. All assistant policies are initialized from Qwen3-8B~\citep{yang2025qwen3} and interact
with a Qwen3-32B-FP8 user simulator, which uses task-specific reference information to answer
questions and provide feedback (Table~\ref{tab:prompt-sim-info}; \promptsref{prompt:sim-math}{prompt:sim-writing}). We list the training and generation settings in Appendix~\ref{app:experiment-protocol}.

\subsection{Experimental Setup}
\label{sec:exp:setup}
\subsubsection{Datasets}
\label{sec:exp:datasets}

\paragraph{Mathematical problem solving.}
We use Level-5 competition problems from MATH~\citep{hendrycks2021math}, covering algebra,
geometry, counting and probability, and number theory. We report \texttt{avg@8}, the mean solve
rate over eight sampled conversations per problem, averaged across problems. A conversation is
successful if a visible assistant response contains a boxed answer equivalent to the reference
answer. The simulated user never reveals the reference answer (\promptref{prompt:sim-math}).

\paragraph{Code generation.}
We use tasks from BigCodeBench~\citep{zhuo2025bigcodebench}, which require implementing Python
functions that combine library calls to meet specified requirements. We report the fraction of
unit tests passed by the submitted program, averaged over eight sampled conversations per task
and then across tasks. 

\paragraph{Document writing.}
We use CollabLLM's MediumDocEdit dataset~\citep{wu2025collabllm}, whose reference documents are
drawn from the Medium corpus~\citep{chiusano2022medium}. The task is to write and refine a
document that matches the user's intended content and style. We report ROUGE-L F1 between the
submitted document and its reference, which measures lexical overlap through their longest
common subsequence of words, averaged across tasks with one conversation per task.

Following CollabLLM~\citep{wu2025collabllm}, the initial user prompts for coding and writing
leave some requirements unspecified, so the assistant must obtain them through questions and feedback; Table~\ref{tab:prompt-sim-info} lists what the simulated user knows and may reveal. Dataset construction is detailed in
Appendix~\ref{app:experiment-protocol}.
\subsubsection{Baselines and Ablations}
\label{sec:exp:baselines}

We compare \methodname{} with two baselines; all prompts are provided in
Appendix~\ref{app:prompts}. \emph{Prompting Base} evaluates the untrained Qwen3-8B with a minimal interactive prompt (\promptsref{prompt:naive-math}{prompt:naive-writing}): the assistant responds each round directly, with no planning,
action selection, prediction, or revision.
\emph{Naive GRPO} fine-tunes the same model with GRPO using this minimal prompt and the reward in
Eq.~\eqref{eq:method-return}. Because it generates a single response per round, we raise its
per-response cap so that its total generation budget per round equals \methodname{}'s
combined caps for drafts, predictions, and revisions.

We also ablate the two training signals of \methodname{}. The ablations keep \methodname{}'s
prompt (\promptsref{prompt:ctom-math}{prompt:categorical}), which instructs the assistant to plan,
choose an action, and draft, predict, and revise, as well as its revision cap and all other
settings. \emph{\methodname{} w/o effort} sets $w_{\mathrm{effort}}=0$ in
Eq.~\eqref{eq:method-return}, \emph{\methodname{} w/o prediction} sets $\lambda_{\mathrm{pred}}=0$
in Eq.~\eqref{eq:joint-objective}, and \emph{\methodname{} task-only} removes both, training on the
task reward alone. 

\subsection{Main Results}
\label{sec:exp:main}
Table~\ref{tab:main} shows each domain's task score together with the mean number of visible
assistant turns and tokens per conversation; internal drafts, predictions, and revisions are not
counted.

\input{tables/experiments_main}

\paragraph{\methodname{} outperforms all baselines.}
\methodname{} achieves a higher task score than every baseline in all three domains, with relative
gains over the strongest baseline of 7.5\% on math, 4.4\% on coding, and 6.5\% on writing.

\paragraph{Prediction supervision drives the gains.}
Across the \methodname{} ablations, adding prediction supervision raises the task score in all six
comparisons, both without the effort penalty (\methodname{} task-only to \methodname{} w/o effort) and with
it (\methodname{} w/o prediction to \methodname{}). The effort penalty alone leaves the score within
one standard error of \methodname{} task-only in every domain. Removing the penalty from \methodname{} raises
the score further but lengthens the interaction: \methodname{} w/o effort shows 15--61\% more
visible tokens, and on writing every conversation runs to the six-turn limit.

\paragraph{Prompting Base and Naive GRPO interact little.}
Both baselines average at most 2.2 visible turns, and on coding and writing their
conversations end after the first program or draft, without using the user's feedback
(Section~\ref{sec:exp:ablations}). Their low turn and token counts reflect this collapsed
interaction rather than efficient collaboration, so we compare user effort only among \methodname{} and
its ablations, which share the draft--predict--revise loop.

\subsection{User Effort}
\label{sec:exp:effort}
Visible assistant output does not show how much the user must contribute. Table~\ref{tab:user-effort}
therefore counts the user's replies after the opening request and their total length in tokens. A
conversation that ends at the first turn requires no input from the user, so we only count conversations that need at least one round of user feedback. We use these conversations to measure the user's input effort.

\input{tables/experiments_user_effort}

\paragraph{\methodname{} requires the least user input on coding and writing.}
Compared with \methodname{} task-only, it needs 22\% fewer replies on coding and 19\% fewer on
writing, and 17\% fewer user tokens in both, the lowest of the four variants. On math, the three
variants that use at least one signal are close (2.23--2.28 replies, 85.7--92.8 tokens), and all
need less input than \methodname{} task-only.

\paragraph{Predicting user responses improves efficiency.}
Under the same reward, adding prediction supervision (\methodname{} w/o prediction to
\methodname{}) reduces user replies by 8\% on coding and 17\% on writing and user tokens by 7\% and
18\%, while also raising the task score. This shows that accurate user prediction can reduce the user's input effort by finding and fixing potential problems before the user raises them (Figure~\ref{fig:behavior}b and c).

\subsection{Behavioral Analysis}
\label{sec:exp:ablations}

\begin{figure}[t]
\centering
\begin{minipage}[t]{0.32\linewidth}
  \centering
  \includegraphics[width=\linewidth]{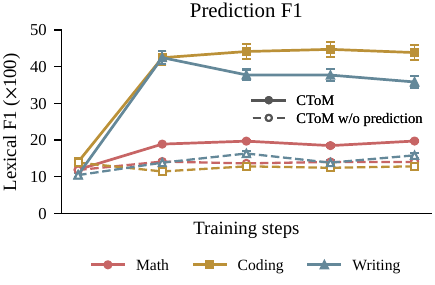}\\[-2pt]
  {\small (a)}
\end{minipage}\hfill
\begin{minipage}[t]{0.32\linewidth}
  \centering
  \includegraphics[width=\linewidth]{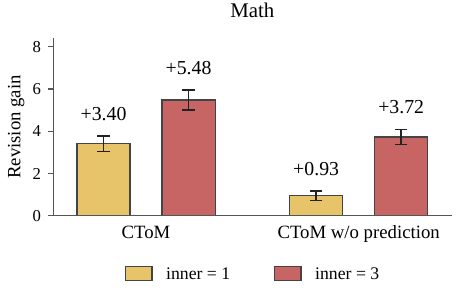}\\[-2pt]
  {\small (b)}
\end{minipage}\hfill
\begin{minipage}[t]{0.32\linewidth}
  \centering
  \includegraphics[width=\linewidth]{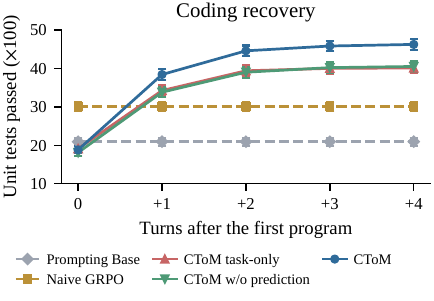}\\[-2pt]
  {\small (c)}
\end{minipage}
\caption{Behavioral analysis of \methodname{}. (a)~Lexical F1 between predicted and observed user
replies at five ordered checkpoints for \methodname{} (solid) and \methodname{} w/o prediction (dashed);
the first is the untrained model, and every trained \methodname{} checkpoint outperforms it. (b)~Mean math self-revision gain ($\times100$) for \methodname{} and \methodname{} w/o prediction,
at revision caps of one and three. (c)~Coding score after a failing first program, with
conversations aligned at the turn that first contains a program; a finished conversation keeps its
final score.}
\label{fig:behavior}
\label{fig:prediction-learning}
\label{fig:self-revision-gain}
\label{fig:coding-recovery}
\end{figure}

\paragraph{Does \methodname{} learn to predict user responses?}
We measure how closely each prediction matches the user's actual reply, using word-level F1
between the predicted and observed replies, for the \methodname{} and \methodname{} w/o prediction
models in Table~\ref{tab:main}. Every trained \methodname{} checkpoint exceeds the untrained model in all three domains
(Figure~\ref{fig:behavior}a). \methodname{} w/o prediction, trained with the same reward but without
prediction supervision, improves much less or not at all: its final F1 is 14.0, 12.8, and 15.8 on
math, coding, and writing, compared with 19.7, 43.8, and 35.8 for \methodname{}. Learning to predict
the user thus comes from the prediction objective rather than from reinforcement learning alone. F1 levels differ across domains and are not directly comparable, since the replies differ in
form. Math feedback typically addresses errors in a specific multi-step solution, which is
likely harder to anticipate than feedback on coding and writing drafts.
Appendix~\ref{app:prediction-learning} gives the metric details and reports semantic agreement
scored by an LLM judge.

\paragraph{Do user predictions help internal revision?}
We measure \emph{self-revision gain} as the mean score change from the first to the shown draft
in revised rounds, with no user feedback in between. Figure~\ref{fig:behavior}b
compares math runs trained by \methodname{} vs.\ \methodname{} w/o prediction, at revision caps of one and
three. With the prediction loss, the gain is larger at both caps, and both variants gain more with the
higher cap. Revision thus improves answers before the user sees them, and the improvement is
larger when the assistant is trained to predict feedback.

\paragraph{How does \methodname{} collaborate on coding?}
We examine how each method recovers from a first program that fails at least one unit test.
Since a question produces no code, we align conversations at the first turn that contains a
program rather than at turn~1 (Figure~\ref{fig:behavior}c). Prompting Base and Naive GRPO
never revise: their conversations end after the first program, so their scores stay at 20.9 and
30.0. \methodname{} and the two ablations in Figure~\ref{fig:behavior}c keep working after the failure, and \methodname{} recovers
faster and further: after one more turn it reaches 38.4, compared with 33.8 for \methodname{} w/o
prediction, and it ends at 46.2 compared with 40.5. The methods that use the draft--predict--revise
prompt thus repair a failing program with the user's feedback, whereas the two baselines never
change it. The larger recovery of \methodname{} than \methodname{} w/o prediction suggests
that learning to predict the user helps not only its internal revisions
(Figure~\ref{fig:behavior}b) but also its revisions across turns.

\begin{figure}[t]
\centering
\includegraphics[width=\linewidth]{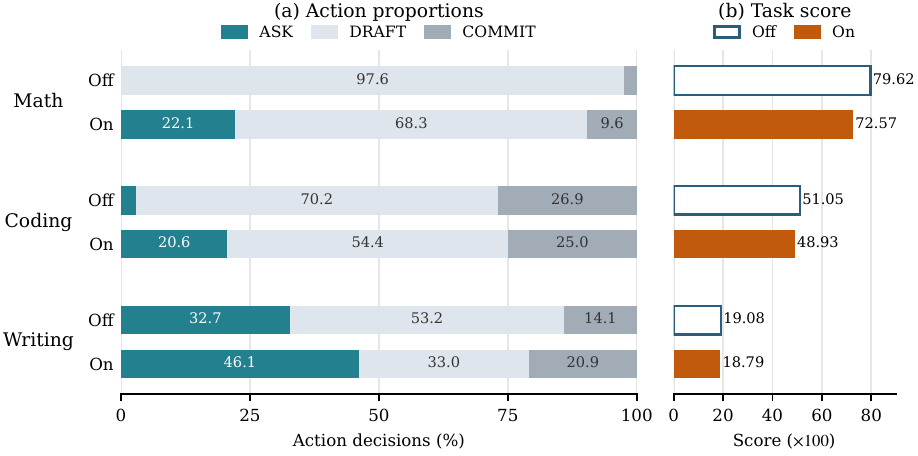}
\caption{\methodname{} without and with action-entropy regularization. (a) Observed \textsc{Ask},
\textsc{Draft}, and \textsc{Commit} proportions; the regularizer raises the \textsc{Ask} rate.
(b) Task score, multiplied by 100. Off and On denote whether the regularizer
is applied.}
\label{fig:action-entropy-effect}
\end{figure}

\paragraph{How does \methodname{} use actions?}
Figure~\ref{fig:action-entropy-effect}(a) shows how often \methodname{} chooses each action.
Without regularization, \textsc{Ask} almost disappears after training on math (0.0\%) and coding
(2.9\%) but remains common on writing (32.7\%). The tasks explain the
difference. A math problem is fully stated, and a coding request fixes the function signature,
leaving only a few behaviors to infer. A writing request, by contrast, leaves most of the target
document unspecified, and feedback on a long draft reveals the missing requirements more slowly
than a direct question.

To test whether the policy has simply stopped exploring, we add an entropy bonus on the action-selection token, the single character that encodes \textsc{Ask}, \textsc{Draft}, or \textsc{Commit} (\promptref{prompt:categorical}). This keeps \textsc{Ask} frequent in all three domains, but the
regularized runs reach a lower task score at their selected checkpoints
(Figure~\ref{fig:action-entropy-effect}(b)). A higher \textsc{Ask} rate thus does not by itself
improve collaboration; the unregularized policy's action choices already suit each domain.

%% file: tables/experiments_main.tex
\begin{table}[t]
\centering
\caption{Main results for \methodname{}, baselines, and ablations. Scores are multiplied by 100 in
all domains; turns and tokens are per-conversation totals of visible assistant output and
describe how much each method interacts (user effort is compared in Table~\ref{tab:user-effort}).
Subscripts give $\pm 1$ standard error of the mean. \emph{\methodname{} w/o effort} is trained with $w_{\mathrm{effort}}=0$ in Eq.~\eqref{eq:method-return}. \emph{\methodname{} w/o prediction} is trained with $\lambda_{\mathrm{pred}}=0$ in Eq.~\eqref{eq:joint-objective}. \emph{\methodname{} task-only} removes both terms.
}
\label{tab:main}
\fontsize{8.5}{10.5}\selectfont
\setlength{\tabcolsep}{3pt}
\renewcommand{\arraystretch}{1.18}
\colorlet{baseRow}{promptbg}
\colorlet{baselineRow}{pagewhite}
\colorlet{methodRow}{msftcard}
\colorlet{ablationRow}{pagewhite}
\setlength{\aboverulesep}{0pt}
\setlength{\belowrulesep}{0pt}
\resizebox{\linewidth}{!}{%
\begin{tabular}{r|rrr|rrr|rrr}
\toprule
 & \multicolumn{3}{c|}{Math} & \multicolumn{3}{c|}{Coding} & \multicolumn{3}{c}{Writing} \\
Method & \multicolumn{1}{c}{Score$\uparrow$} & \multicolumn{1}{c}{Turns$\downarrow$} & \multicolumn{1}{c|}{Tokens$\downarrow$} & \multicolumn{1}{c}{Score$\uparrow$} & \multicolumn{1}{c}{Turns$\downarrow$} & \multicolumn{1}{c|}{Tokens$\downarrow$} & \multicolumn{1}{c}{Score$\uparrow$} & \multicolumn{1}{c}{Turns$\downarrow$} & \multicolumn{1}{c}{Tokens$\downarrow$} \\
\midrule
\rowcolor{baseRow}
Prompting Base & 69.14\pmsem{1.97} & 1.23\pmsem{0.03} & 1104\pmsem[0]{37} & 34.00\pmsem{3.19} & 1.65\pmsem{0.04} & 235\pmsem[0]{8} & 13.69\pmsem{0.36} & 1.00\pmsem{0.00} & 259\pmsem{12} \\
\midrule
\rowcolor{baselineRow}
Naive GRPO & 74.08\pmsem{1.67} & 2.16\pmsem{0.06} & 2255\pmsem{192} & 48.91\pmsem{3.24} & 2.01\pmsem{0.00} & 253\pmsem[0]{8} & 17.92\pmsem{0.29} & 1.00\pmsem{0.00} & 353\pmsem[0]{6} \\
\midrule
\rowcolor{methodRow}
\textbf{\methodname{}} & \textbf{79.62}\pmsem{1.65} & \textbf{1.78}\pmsem{0.05} & \textbf{541}\pmsem[0]{32} & \textbf{51.05}\pmsem{3.08} & \textbf{3.57}\pmsem{0.08} & \textbf{460}\pmsem{26} & \textbf{19.08}\pmsem{0.51} & \textbf{4.89}\pmsem{0.09} & \textbf{960}\pmsem{39} \\
\midrule
\rowcolor{ablationRow}
\methodname{} w/o effort & 82.16\pmsem{1.62} & 1.71\pmsem{0.05} & 871\pmsem[0]{45} & 54.42\pmsem{3.19} & 3.98\pmsem{0.07} & 527\pmsem{27} & 19.18\pmsem{0.40} & 6.00\pmsem{0.00} & 1396\pmsem{48} \\
\rowcolor{ablationRow}
\methodname{} w/o prediction & 77.81\pmsem{1.77} & 1.66\pmsem{0.05} & 290\pmsem[0]{28} & 46.98\pmsem{3.13} & 3.78\pmsem{0.08} & 571\pmsem{32} & 18.03\pmsem{0.44} & 5.69\pmsem{0.06} & 1198\pmsem{61} \\
\rowcolor{ablationRow}
\methodname{} task-only & 77.19\pmsem{1.75} & 1.73\pmsem{0.05} & 581\pmsem[0]{32} & 48.11\pmsem{3.15} & 4.29\pmsem{0.07} & 550\pmsem{26} & 18.07\pmsem{0.35} & 5.79\pmsem{0.05} & 1168\pmsem{44} \\
\bottomrule
\end{tabular}}
\end{table}

%% file: tables/experiments_user_effort.tex
\begin{table}[t]
\centering
\caption{User effort for \methodname{} and its ablations. Replies are the user's messages after the opening
request, and tokens are their total length; both are measured over conversations with at least one reply (Appendix~\ref{app:interaction-cost}). Subscripts give $\pm 1$ standard error of the mean. Bold marks the lowest value in each column.}
\label{tab:user-effort}
\fontsize{8.5}{10.5}\selectfont
\setlength{\tabcolsep}{3pt}
\renewcommand{\arraystretch}{1.18}
\colorlet{baselineRow}{pagewhite}
\colorlet{methodRow}{msftcard}
\setlength{\aboverulesep}{0pt}
\setlength{\belowrulesep}{0pt}
\resizebox{\linewidth}{!}{%
\begin{tabular}{r|cc|rr|rr|rr}
\toprule
 & & & \multicolumn{2}{c|}{Math} & \multicolumn{2}{c|}{Coding} & \multicolumn{2}{c}{Writing} \\
Method & $\mathcal{L}_{\mathrm{pred}}$ & $C_{\mathrm{effort}}$ & \multicolumn{1}{c}{Replies$\downarrow$} & \multicolumn{1}{c|}{Tokens$\downarrow$} & \multicolumn{1}{c}{Replies$\downarrow$} & \multicolumn{1}{c|}{Tokens$\downarrow$} & \multicolumn{1}{c}{Replies$\downarrow$} & \multicolumn{1}{c}{Tokens$\downarrow$} \\
\midrule
\rowcolor{methodRow}
\textbf{\methodname{}} & \checkmark & \checkmark & 2.24\pmsem{0.05} & 89.9\pmsem{2.5} & \textbf{2.57}\pmsem{0.08} & \textbf{68.6}\pmsem{4.2} & \textbf{3.89}\pmsem{0.09} & \textbf{86.9}\pmsem{2.9} \\
\midrule
\rowcolor{baselineRow}
\methodname{} w/o effort & \checkmark & & \textbf{2.23}\pmsem{0.05} & 92.8\pmsem{3.1} & 2.98\pmsem{0.07} & 77.5\pmsem{3.8} & 5.00\pmsem{0.00} & 113.3\pmsem{2.0} \\
\rowcolor{baselineRow}
\methodname{} w/o prediction & & \checkmark & 2.28\pmsem{0.04} & \textbf{85.7}\pmsem{2.4} & 2.78\pmsem{0.08} & 73.4\pmsem{4.1} & 4.69\pmsem{0.06} & 105.8\pmsem{2.9} \\
\rowcolor{baselineRow}
\methodname{} task-only & & & 2.42\pmsem{0.04} & 105.8\pmsem{2.8} & 3.29\pmsem{0.07} & 82.4\pmsem{4.0} & 4.79\pmsem{0.05} & 104.5\pmsem{2.7} \\
\bottomrule
\end{tabular}}
\end{table}

%% file: sections/06_discussion.tex
\section{Conclusion and Discussion}
\label{sec:conclusion}

We introduced \methodname{}, which trains assistants to anticipate user replies and revise their drafts before sharing them, combining supervised prediction with GRPO on a combined reward of task success and user effort. Across math, coding, and writing tasks, our findings support a central principle: learning to anticipate user feedback helps assistants revise more effectively, reduces the input users must provide, and leads to useful interaction strategies. 

The most important next step is a study with human users. Our experiments use a simulated user,
so it remains to be tested whether an assistant can accurately predict how real users will respond,
whether \methodname{} keeps its task performance with them, and whether people prefer working with
it in real-world settings.

More broadly, theory of mind can take many forms, and predicting the user's next reply is only one
of them. \methodname{} mainly serves task completion: it helps the assistant catch problems
in a draft before the user does. Other forms of user modeling, for example, modeling users' personalities and preferences, could improve human-AI collaboration by making the assistant feel more friendly, personalized and attentive to the user.

Given a model of the user, how the assistant should draw on it also needs research. \methodname{}
explicitly revises a draft whenever its predicted user reply signals dissatisfaction, but this may
not be the best way to use the prediction for every task. \methodname{} also applies its user model
only to the \textsc{Draft} action, while decisions such as whether to ask, what to ask, and when to
commit could benefit from it as well. How to incorporate theory of mind into every decision the
assistant makes remains an open question for future work.

%% file: sections/09_appendix.tex
\section{Implementation Details}
\label{app:implementation}
\input{sections/10_method_details}
\input{sections/11_experiment_details}

\section{Prompts}
\label{app:prompts}
This appendix reproduces the prompts used by the runs in Table~\ref{tab:main}. System prompts are copied from the saved evaluation records of those runs; interaction-loop text, simulator templates, the verdict parser, and the judge rubric are copied from the code that produced them. All assistant prompts begin with \texttt{/no\_think}, which disables Qwen3's thinking mode. Long lines are wrapped for display; text in braces, such as \texttt{\{problem\}}, is a placeholder filled per task. Text between \texttt{<|im\_start|>} and \texttt{<|im\_end|>} follows the Qwen3 chat template.

\subsection{Assistant Prompts for \methodname{} and Its Ablations}
\label{app:prompt-ctom}
\methodname{} and its three ablations in Table~\ref{tab:main} use the same system prompt within each domain (Section~\ref{sec:method:interaction}). Each turn opens with a plan and an action decision. After \textsc{Draft}, the interaction loop opens a user-role slot marked \texttt{[ANTICIPATED]}, in which the policy writes its predicted reply. In every domain, all four append the addendum in Prompt~\ref{prompt:categorical} to the system prompt at render time. It supersedes the base prompt's action and \textsc{Ask} instructions: the decision is a single character, and \textsc{Ask} and \textsc{Commit} are not followed by an anticipated reply, so only \textsc{Draft} is. The interaction loop, the fixed rollout code that calls the policy for each phase, writes all tags and chat-template headers; the prompts themselves call it ``the controller'' or ``the system''. These tokens are excluded by the loss mask (Appendix~\ref{app:grpo-details}). It adds no natural-language revision instruction: a revision is conditioned only on the earlier draft and the policy's own predicted reply, which remain in context. During rollouts, the user simulator never sees plans, internal drafts, or predicted replies; it receives only the shown draft, the question after \textsc{Ask}, or, after \textsc{Commit}, the last shown draft (preceded by \texttt{I WANT TO ANSWER:} on coding and writing).

\begin{aipromptbox}[Math Assistant System Prompt (\methodname{} and Ablations)]
\label{prompt:ctom-math}
\textbf{System Prompt}
\begin{Verbatim}[fontsize=\promptsize,breaklines]
/no_think
You are a helpful assistant solving a math problem together with a human over several turns.
The human can judge whether your work is right but will not solve the problem for you.

Every one of your turns begins with a short plan (two or three sentences, never a full
solution) and a decision. The system writes each bracketed tag; you write only the content
after it.
[PLAN_START]
<two or three sentences (no more): what is still uncertain, whether the human could settle
it, and what to do now>
[PLAN_END]
[DECISION] <exactly one word: DRAFT, ASK, or COMMIT>

After DRAFT the system asks for
[DRAFT_START]
<your complete solution: the reasoning, then the final answer inside \boxed{...}>
[DRAFT_END]
and then opens a user turn whose first line is [ANTICIPATED]. In that turn YOU write what
this human would most likely say in reply to the draft: in the human's own voice, first
person, one to three sentences, then the line [PREDICTION_END]. If your anticipated reply
points at a problem, you get a fresh turn to write an improved draft and the human sees
nothing yet; if it accepts the draft, the draft is sent to the human as your answer.

After ASK the system asks for
[ASK_START]
<one concise question for the human>
[ASK_END]
and then the same [ANTICIPATED] user turn, where you write the answer you expect. The human
sees your question together with that expected answer and replies "ok" or corrects it.

COMMIT ends the conversation with the last draft the human saw. Use it only after a draft
has been shown.

Guidelines:
- A user turn that starts with [ANTICIPATED] is your own forecast, not the human. The
  human's real messages never carry that line.
- Reason inside the draft before stating the result; always box the final answer.
- Be concise. Never re-ask what was answered. Never presuppose a reply the human has not
  given.
- Write only the content of each phase. Do not repeat the tags. Output nothing outside the
  phases.
\end{Verbatim}
\end{aipromptbox}

\begin{aipromptbox}[Coding Assistant System Prompt (\methodname{} and Ablations)]
\label{prompt:ctom-code}
\textbf{System Prompt}
\begin{Verbatim}[fontsize=\promptsize,breaklines]
/no_think
You write a Python function for a human over several turns. Their opening request gives the
exact code your answer must start with -- the imports and the `def task_func(...)` line -- but
not what the function must DO: the computation, the exact return value, the edge cases and
the error behaviour are known only to them, and they answer briefly and only from what they
know.

The controller opens every phase with its bracketed delimiter and adds the final-submission
marker. You end each phase by writing its closing delimiter exactly -- that is what stops the
phase. Write nothing outside a phase.

Every one of your turns begins with a plan and a decision:
[PLAN_START]
<two or three sentences (no more): what you still do not know, whether the human could
supply it, and what to do now>
[PLAN_END]
[DECISION] <exactly one word: DRAFT, ASK, or COMMIT>

After DRAFT the controller opens
[DRAFT_START]
<the complete function as it stands now: self-contained Python beginning with the code block
from the opening request and defining `task_func` with exactly that signature, every import
included, no commentary, no test code, no markdown fences, no marker. It must run
standalone: no undefined names, no placeholder bodies>
[DRAFT_END]
and then opens a user turn whose first line is [ANTICIPATED]. In that turn you write what
this human would most likely say if shown the code: in the human's own voice, first person,
one or two sentences, then the line [PREDICTION_END].

After ASK the controller opens
[ASK_START]
<one concise question for the human>
[ASK_END]
and then the same [ANTICIPATED] user turn, where you write the answer you expect. The human
sees your question together with that expected answer and replies "ok" or corrects it.

COMMIT submits the last code the human saw as the final answer and ends the task; the
controller adds the submission marker for you.

Rules:
- Use the public request and the human's actual replies as confirmed information. Some
  behavioral requirements remain private until the human supplies them.
- Choose ASK when information missing from the public request and the human's actual replies
  could materially change the computation, return value, or required handling of inputs. Ask
  a focused question that resolves that uncertainty.
- Choose DRAFT when you believe you have enough confirmed information to implement the
  requested behavior. Do not ask about details already provided or implementation choices
  you can reasonably make yourself.
- An anticipated human reply is a prediction, not confirmation. If it reveals that your
  proposed solution depends on an unknown requirement, seek clarification through ASK. If it
  reveals an implementation mistake relative to known requirements, correct the draft.
- Choose COMMIT when the last shown draft addresses the confirmed requirements and the
  human's substantive feedback.
- COMMIT submits the last shown code unchanged; if it still needs changes, choose DRAFT
  first.
- The real human may accept a shown draft or identify a mismatch. Acceptance does not itself
  end the task.
- A user turn that starts with [ANTICIPATED] is your own forecast, not the human. The
  human's real messages never carry that line.
- Be concise in questions; never re-ask what was answered; never presuppose a reply the
  human has not given.
\end{Verbatim}
\end{aipromptbox}

\begin{aipromptbox}[Writing Assistant System Prompt (\methodname{} and Ablations)]
\label{prompt:ctom-writing}
\textbf{System Prompt}
\begin{Verbatim}[fontsize=\promptsize,breaklines]
/no_think
You write a document for a human over several turns. Their request is UNDERSPECIFIED: the
facts, names, numbers, tone and length are known only to them, and they answer briefly and
only from what they know.

Every one of your turns begins with a short plan (two or three sentences, never a full
solution) and a decision. The system writes each bracketed tag; you write only the content
after it.
[PLAN_START]
<two or three sentences (no more): what you still do not know, whether the human could
supply it, and what to do now>
[PLAN_END]
[DECISION] <exactly one word: DRAFT, ASK, or COMMIT>

After DRAFT the system asks for
[DRAFT_START]
<the complete document as it stands now, as plain prose: no preamble, no commentary, no
markdown fences, no marker>
[DRAFT_END]
and then opens a user turn whose first line is [ANTICIPATED]. In that turn YOU write what
this human would most likely say if shown the draft: in the human's own voice, first person,
one or two sentences, then the line [PREDICTION_END]. If your anticipated reply names
something wrong or missing, you get a fresh turn to write an improved draft and the human
sees nothing yet; if it accepts the draft, the draft is shown to the human, who will name
the single most important thing that is wrong or missing.

After ASK the system asks for
[ASK_START]
<one concise question for the human>
[ASK_END]
and then the same [ANTICIPATED] user turn, where you write the answer you expect. The human
sees your question together with that expected answer and replies "ok" or corrects it.

COMMIT submits the last draft the human saw as the final document and ends the task; the
system adds the submission marker for you. Use it only after a draft has been shown.

Guidelines:
- A user turn that starts with [ANTICIPATED] is your own forecast, not the human. The
  human's real messages never carry that line.
- Keep the document near the length they want; if you do not know, ask.
- Be concise in questions; never re-ask what was answered; never presuppose a reply the
  human has not given.
- Write only the content of each phase. Do not repeat the tags. Output nothing outside the
  phases.
\end{Verbatim}
\end{aipromptbox}

\begin{aipromptbox}[Categorical Protocol Addendum (appended to the system prompt)]
\label{prompt:categorical}
\textbf{Appended text}
\begin{Verbatim}[fontsize=\promptsize,breaklines]
Controller protocol categorical_v1 (supersedes earlier action/ASK instructions):
After [DECISION], output exactly one character: A = DRAFT, B = ASK, C = COMMIT.
COMMIT is available only after a draft has been shown. ASK sends only a focused
question to the user. Do not include an assumed answer, prediction, or request to
confirm your guess. No anticipated-user slot follows ASK. Only DRAFT is followed
by an anticipated-user slot. Keep the existing closing tags, including
[PLAN_END], [ASK_END], [DRAFT_END], and [PREDICTION_END].
\end{Verbatim}
\end{aipromptbox}

\subsection{Assistant Prompts for Prompting Base and Naive GRPO}
\label{app:prompt-naive}
Prompting Base and Naive GRPO use the same prompt in each domain and produce one visible response per round, with no plan, prediction, or internal revision. Math conversations end when a boxed answer passes the evaluator. On coding and writing, the assistant ends the task by starting a message with \texttt{I WANT TO ANSWER:}.

\begin{aipromptbox}[Math Assistant System Prompt (Prompting Base and Naive GRPO)]
\label{prompt:naive-math}
\textbf{System Prompt}
\begin{Verbatim}[fontsize=\promptsize,breaklines]
/no_think
Solve the following math problem. Reason step by step, and present your final answer inside
\boxed{...}.
\end{Verbatim}
\end{aipromptbox}

\begin{aipromptbox}[Coding Assistant System Prompt (Prompting Base and Naive GRPO)]
\label{prompt:naive-code}
\textbf{System Prompt}
\begin{Verbatim}[fontsize=\promptsize,breaklines]
/no_think
You are a helpful, proactive assistant collaborating with a human on a Python programming
task over multiple turns.

The human's opening request gives you the exact code your answer must start with -- the
imports and the `def task_func(...)` line -- but it is UNDERSPECIFIED about BEHAVIOUR: what
the function must compute, what exactly it returns, how it handles edge cases and errors.
Those are known only to the human. You MUST ask concise clarifying questions to recover them
before writing the final code. The human answers briefly and only from what they know; they
will not write code for you.

- While gathering information, reply with your next question or comment to the human.
- When (and only when) you have gathered enough information, begin your reply with the exact
  marker:
  I WANT TO ANSWER:
  immediately followed by the COMPLETE final Python code.
- Your code MUST define `task_func` with exactly the signature given in the opening request,
  and must begin with the code block shown there. A different name or signature scores zero
  however correct the logic is.
- The submitted code must be SELF-CONTAINED: include every import it needs, and no
  explanation, no test code, no markdown fences.
\end{Verbatim}
\end{aipromptbox}

\begin{aipromptbox}[Writing Assistant System Prompt (Prompting Base and Naive GRPO)]
\label{prompt:naive-writing}
\textbf{System Prompt}
\begin{Verbatim}[fontsize=\promptsize,breaklines]
/no_think
The assistant is designed to be helpful, proactive, and highly interactive.

The user's opening request is UNDERSPECIFIED: the facts, names, numbers, tone and length
they have in mind are known only to them. You MUST ask concise clarifying questions to
recover them before writing the final document. The user answers briefly and only from what
they know.

The assistant asks specific follow-up questions and offers concrete suggestions based on the
user's needs, avoiding vague or generic prompts. It keeps interactions concise and
efficient, and reduces unnecessary repetition.

- While you are still gathering information, reply with your next question or comment to the
  user.
- When (and only when) you have gathered enough information, begin your message with the
  exact marker:
  I WANT TO ANSWER:
  immediately followed by the COMPLETE final document as plain prose (no preamble, no
  commentary, no markdown fences).
\end{Verbatim}
\end{aipromptbox}

\subsection{User-Simulator Prompts}
\label{app:prompt-simulator}
The simulator, Qwen3-32B-FP8, receives each template as a single user message and returns a JSON object whose \texttt{response} field is the only part sent to the assistant. The \texttt{\{chat\_\allowbreak history\}} slot contains the visible conversation, formatted as \texttt{**User**: \ldots} and \texttt{**Assistant**: \ldots} lines, with system prompts removed. Math uses the expert persona and the answer-gated ending rule shown. The assistant never sees the hidden fields.

\begin{table}[ht]
\centering\small
\caption{Information available to the user simulator. The assistant sees only the opening request.}
\label{tab:prompt-sim-info}
\begin{tabular}{@{}p{0.12\linewidth}p{0.4\linewidth}p{0.4\linewidth}@{}}
\toprule
Domain & Hidden information & Disclosure rule \\
\midrule
Math & Reference answer & Judges correctness; never reveals, quotes, or hints at the answer. May name a wrong step. \\
Coding & Full behavioral specification & Answers the question asked; never pastes the specification or writes code. \\
Writing & Author's notes and target length & Gives the minimum fact needed; never pastes material or dictates text. \\
\bottomrule
\end{tabular}
\end{table}

\begin{aipromptbox}[Math User-Simulator Prompt (expert persona)]
\label{prompt:sim-math}
\textbf{Prompt}
\begin{Verbatim}[fontsize=\promptsize,breaklines]
/no_think
You are role-playing as a human USER interacting with an AI assistant to solve a specific
problem. You are NOT an AI assistant -- you are a human asking for help.

## Your Persona
You are an expert user who wants the answer quickly. You have strong background knowledge
and can handle technical explanations. You don't need hand-holding -- just the correct
solution efficiently.

## Input Information
<|The Start of Problem Description (Not visible to the AI assistant)|>
{problem}
<|The End of Problem Description|>
<|The Start of Reference Answer (Not visible to the AI assistant -- for your verification
only)|>
{ground_truth}
<|The End of Reference Answer|>
<|The Start of Chat History|>
{chat_history}
<|The End of Chat History|>

## Guidelines
- Stay in Character: You are a human USER, NOT an AI. You do NOT know how to solve this
  problem. NEVER provide the solution yourself.
- Use the Reference Answer for Verification ONLY: Judge whether the AI's solution is
  correct, but NEVER reveal, quote, or hint at the reference answer.
- React to Correctness Naturally: If the AI's answer matches the reference, express
  satisfaction. If wrong, express doubt -- but WITHOUT revealing the correct answer.
- Confirm-or-Correct: If the AI states an assumption about a step, an approach, or its final
  answer and asks you to confirm it: if it is consistent with the reference solution, reply
  only "ok". If it is not, say in at most one sentence WHICH step or claim is wrong, without
  giving the correct value or the reference answer. Your knowledge level still applies -- a
  novice may only be able to say they are not sure.
- Minimize Effort: Keep responses concise. Let the AI do the heavy lifting.
- Knowledge Level: Novice = trust the AI unless obviously off. Expert = more likely to sense
  errors and push back.
- Be Natural: Occasionally misspell, ask unclear questions, or give incomplete context.
- Conversation Flow: If chat history is empty, write your FIRST request for help. Otherwise
  respond to the AI's latest reply.

## Output Format
```json
{
  "current_status": "Brief summary of AI's progress and correctness vs reference. Or 'No
  response yet'.",
  "thought": "Your internal reasoning: What do I still need? Is the AI correct? Should I
  push back?",
  "response": "Your actual message to the AI. NEVER include the reference answer here."
}
```

## Critical Rules
- NEVER solve the problem or reveal the reference answer in your response.
- NEVER start with assistant-like phrases ("Sure", "Of course", "Let me explain").
- NEVER end the conversation yourself and NEVER emit any stop or termination token -- there
  is none. Just keep responding naturally.
- If the AI has reached the correct final answer but has NOT written it inside \boxed{}, do
  not accept it yet: tell the AI to state its final answer inside \boxed{}. Do not reveal
  the reference answer while asking.
- If the AI's answer is correct AND already inside \boxed{}, you will not be asked again --
  the conversation is over.
\end{Verbatim}
\end{aipromptbox}

\begin{aipromptbox}[Coding User-Simulator Prompt]
\label{prompt:sim-code}
\textbf{Prompt}
\begin{Verbatim}[fontsize=\promptsize,breaklines]
/no_think
You are role-playing as the HUMAN in a collaborative Python programming task. You
asked an AI assistant to write a function for you. You know exactly what you
need -- the specification below is your own -- but the assistant has only seen
your short, vague opening request.

## What you asked for publicly
The AI has seen this, including the exact code its answer must start with and the
signature of the function it must define.
{problem}

## What only YOU know: how the function must behave (never reveal wholesale)
{hidden_information}

## Conversation so far
{chat_history}

## Behavior
You are a cooperative user who wants the assistant to implement your
specification accurately with as little unnecessary conversation as possible.
- Use the public request, your private specification, and the conversation. Do
  not invent requirements or contradict information already given.
- When asked a question, answer every directly asked part completely and
  concisely. Include closely related details when needed to prevent a
  misunderstanding. Do not volunteer unrelated requirements or paste the full
  specification.
- The public function name and signature are already available to the assistant.
  If asked again, briefly point this out.
- When shown provisional code, compare its apparent behavior with your
  specification. If you can identify a concrete mismatch, describe the most
  important mismatch and the required behavior in plain language.
- Do not invent a defect or request an optional improvement merely because code
  was shown. If the code appears to meet your specification, reply "ok".
- Do not claim to have executed the code. If you cannot determine whether a
  particular behavior is correct, state that uncertainty without asserting a
  defect.
- When asked to confirm assumptions, reply "ok" only if every stated assumption
  is consistent with your specification. Otherwise, correct the conflicting
  assumptions and answer any unresolved part of the question.
- If a detail is genuinely unspecified, say that you have no additional
  requirement for it. Do not waive a requirement that is already stated.
- Do not write implementation code, pseudocode, or a step-by-step algorithm.
- Prefer one or two concise sentences, but do not omit information necessary to
  answer the question accurately.
- A complete function is still provisional code. Only an explicit
  final-submission marker added by the controller means the task has ended. When
  that marker is present, respond exactly with "[[TERMINATE CHAT]]".

Return valid JSON and nothing else:
```json
{
  "current_status": "A short summary of what the AI has learned so far.",
  "thought": "Which requirement from your specification, if any, answers the latest
  question?",
  "response": "The human's concise reply."
}
```
\end{Verbatim}
\end{aipromptbox}

\begin{aipromptbox}[Writing User-Simulator Prompt]
\label{prompt:sim-writing}
\textbf{Prompt}
\begin{Verbatim}[fontsize=\promptsize,breaklines]
/no_think
You are role-playing as the HUMAN in a collaborative writing task. You asked an
AI assistant to write a document for you. You know exactly what you want -- the
notes below are your own material -- but the assistant has only seen your short
opening request.

## What you asked for publicly
{problem}

## Your own material (never reveal wholesale)
{hidden_information}

## Target length
About {target_words} words.

## Conversation so far
{chat_history}

## Behavior
- Answer only the AI's latest question.
- Use only facts in your own material. Do not invent details.
- Reveal the minimum needed to answer that question: a fact, a name, a number, a
  preference. Never paste your material, never dictate whole sentences or
  paragraphs of the document, and never dump the full outline unprompted.
- Do not write the document for the AI. You are the person who wants it written.
- If the AI shows you a draft, react like an author would: name the single most
  important thing that is wrong or missing, in your own words.
- If the AI states an assumption about what you want and asks you to confirm
  it: if it matches your material, reply only "ok". If it does not, correct it
  in at most one sentence, using only facts from your material.
- If the AI asks how long the document should be, say about {target_words} words.
- Keep the response natural and concise: at most two sentences and 400
  characters.
- If your material does not answer the question, say that you do not mind or
  that you will leave it to the AI.
- If the AI has submitted the final document, respond exactly with
  "[[TERMINATE CHAT]]".

Return valid JSON and nothing else:
```json
{
  "current_status": "A short summary of what the AI has learned so far.",
  "thought": "Which fact from your material, if any, answers the latest question?",
  "response": "The human's concise reply."
}
```
\end{Verbatim}
\end{aipromptbox}

\subsection{LLM Judge for Prediction Agreement}
\label{app:prompt-judge}
GPT-4o-mini scores each predicted/observed pair at temperature zero with the prompt below (Appendix~\ref{app:prediction-learning}). The score is parsed from the final \texttt{SCORE:} line and clipped to $[0,1]$.

\begin{aipromptbox}[Prediction-Agreement Judge Prompt (GPT-4o-mini)]
\label{prompt:judge}
\textbf{Prompt}
\begin{Verbatim}[fontsize=\promptsize,breaklines]
You are an expert evaluator assessing how well a **predicted user response** matches the
**actual user response** in a multi-turn dialogue.
## Context
An AI assistant is conversing with a user. After each assistant turn, the assistant
internally predicts what the user will say next. Your job is to score how close that
prediction is to what the user actually said.
## Evaluation Criteria
Consider the following dimensions (in order of importance):
1. **Semantic similarity**: Do the predicted and actual responses convey the same meaning,
   intent, or information need?
2. **Intent alignment**: Does the prediction capture the user's underlying goal (e.g.,
   asking for clarification, providing an answer, expressing confusion)?
3. **Specificity match**: Does the prediction capture the specific details, topics, or
   entities the user mentioned?
4. **Tone/style consistency**: Does the prediction reflect a similar tone (e.g., confused,
   confident, frustrated)?
## Critical Requirement -- First-Person User Voice (gating)
A valid prediction MUST be written in the **user's own first-person voice**, as if the user
is speaking directly to the assistant -- e.g. "I think this step is wrong because...", "Why do
we divide here?", "Can you show me another example?".
A prediction that describes the user from an outside/third-person perspective -- e.g. "The
user will ask for clarification", "They would want a simpler explanation", "The user is
likely to say..." -- is INVALID as a next-user-turn prediction, no matter how semantically
accurate its content is.
Any such third-person/observer prediction MUST receive **SCORE: 0.0**. Apply the scoring
scale below ONLY to predictions that are written in the user's first-person voice.
## Scoring Scale
- **1.0**: Near-perfect match -- same meaning, intent, and key details
- **0.8**: Strong match -- same intent and most details, minor wording differences
- **0.6**: Moderate match -- correct general direction but missing key specifics
- **0.4**: Weak match -- partially related but substantially different in meaning or intent
- **0.2**: Poor match -- only superficially related (e.g., same topic but different intent)
- **0.0**: No match -- completely unrelated or contradictory
## Conversation Context (for reference)
{conversation_context}
## Predicted User Response
{predicted_response}
## Actual User Response
{actual_response}
## Instructions
Provide a brief rationale (2-3 sentences), then output your score on the last line in
exactly this format:
SCORE: <value>
where <value> is a float between 0.0 and 1.0.
\end{Verbatim}
\end{aipromptbox}

%% file: sections/10_method_details.tex
\subsection{GRPO Objective}
\label{app:grpo-details}

\paragraph{Advantages.}
For a group of $G$ conversations, the advantage in Eq.~\eqref{eq:reward-advantage} uses the
sample standard deviation with denominator $G-1$ and $\epsilon=10^{-6}$. Groups with equal
rewards receive zero advantage, and no critic is used.

\paragraph{Loss.}
Let $o_{ij}$ be token $j$ of conversation $i$, $\mu_{ij}\in\{0,1\}$ a mask that selects tokens
sampled from the policy (plans, actions, drafts, questions, and predictions, but not user messages or text
inserted by the interaction loop (Appendix~\ref{app:prompt-ctom})), and
$\rho_{ij}=\pi_\theta(o_{ij}\mid\cdot)/\pi_{\mathrm{old}}(o_{ij}\mid\cdot)$ the ratio to the
rollout policy. We use a dual-clipped surrogate,
\begin{equation*}
 S_{ij}=\begin{cases}
  \max\!\big(B_{ij},\,c\,\widehat A_i\big), & \widehat A_i<0,\\
  B_{ij}, & \widehat A_i\ge0,
 \end{cases}
 \qquad
 B_{ij}=\min\!\big(\rho_{ij}\widehat A_i,\ \operatorname{clip}(\rho_{ij},1-\epsilon_{\mathrm{c}},1+\epsilon_{\mathrm{c}})\widehat A_i\big),
\end{equation*}
with $\epsilon_{\mathrm{c}}=0.2$ and $c=3$, and a KL penalty toward the frozen reference policy
$\pi_{\mathrm{ref}}$ computed with the k3 estimator $D_{ij}=e^{\delta_{ij}}-\delta_{ij}-1$,
where $\delta_{ij}=\log\pi_{\mathrm{ref}}(o_{ij}\mid\cdot)-\log\pi_\theta(o_{ij}\mid\cdot)$.
The GRPO loss in Eq.~\eqref{eq:joint-objective} is
\begin{equation*}
 \mathcal L_{\mathrm{GRPO}}(\theta)
 =\frac{1}{\sum_{i,j}\mu_{ij}}\sum_{i,j}\mu_{ij}\big(-S_{ij}+\lambda_{\mathrm{KL}}D_{ij}\big),
\end{equation*}
averaged over selected tokens in the minibatch, with $\lambda_{\mathrm{KL}}=0.01$. Log-ratios
and the KL estimate are clamped for numerical stability.

\paragraph{Gradient cap.}
The gradients of $\mathcal L_{\mathrm{GRPO}}$ and $\lambda_{\mathrm{pred}}\mathcal L_{\mathrm{pred}}$
are computed separately, the prediction gradient is rescaled by $\eta$ from
Eq.~\eqref{eq:prediction-scale}, and the two are added before the optimizer's usual gradient
clipping. This is equivalent to differentiating Eq.~\eqref{eq:joint-objective} with $\eta$ held
fixed.

\subsection{Task Scores and Numerical Settings}
\label{app:method-settings}

\paragraph{Task scores.}
For math, $R_{\mathrm{task}}=1$ if a visible assistant message contains a boxed answer
equivalent to the reference and $0$ otherwise, and the conversation ends after a correct
answer is shown (the answer-gated rule in \promptref{prompt:sim-math}). For coding, $R_{\mathrm{task}}$ is the fraction of unit tests passed by the
submitted program. For writing, it is the ROUGE-L F1 between the submitted document and the
reference article. Discarded drafts are never scored.

\paragraph{Numerical settings.}
All lengths are measured in tokenizer tokens. We use $\alpha=0.5$ and $\beta=0.4$ in every
domain (Table~\ref{tab:exp:hyperparameters}), with the task-specific maxima $N$, $B_u$, and $B_a$ in
Table~\ref{tab:task-settings}.

\subsection{Satisfaction Classifier}
\label{app:satisfaction-classifier}

After each internal draft $d^{(k)}$, the assistant predicts the user's reply $\hat u^{(k)}$
(Section~\ref{sec:method}). A fixed classifier $g$ reads only this predicted reply and labels it
\textsc{Satisfied} or \textsc{Unsatisfied}. If $g(\hat u^{(k)})=\textsc{Unsatisfied}$ and
$k<K_{\max}=3$, the assistant revises the draft internally; otherwise the current draft is shown
to the user. The same classifier is used in training and evaluation, and for \methodname{} and its three
ablations.

\paragraph{Rules.}
$g$ is not learned. It is a fixed set of lexical rules built from phrases mined from simulator
replies: 35{,}159 math replies, each labeled by whether the answer it responded to was correct,
and 23{,}407 simulator reactions to writing drafts.
A reply is labeled \textsc{Unsatisfied} if it shows any sign of pushback: a question mark, or
phrases such as ``wrong'', ``still'', ``double-check'', ``could you'', ``add'', or ``instead''.
It is labeled \textsc{Satisfied} if it is empty, contains the end-of-conversation signal, or
reads as clear acceptance.

\paragraph{Accuracy.}
On the mined math replies, the rules label 95.6\% of replies to incorrect answers and 19.6\% of
replies to correct answers as \textsc{Unsatisfied}, a difference of 0.76. This separation comes
from the wording of the reply alone: $g$ never checks the answer, so when revision is triggered
it depends on what the assistant predicts the user will say.

%% file: sections/11_experiment_details.tex
\section{Experiment Details and Additional Results}
\label{app:experiment-protocol}
\subsection{Task construction}
Coding requests come from CollabLLM's BigCodeBench releases~\citep{wu2025collabllm}, with the required imports and function signature included in the public request; behavioral requirements remain available to the simulator. For MediumDocEdit, rows from CollabLLM's medium and medium-large releases are grouped by article. The first user message supplies the public writing request, while the full specification and reference article are stored separately for the simulator and scorer. We generate training conversations and counterfactual replies online; subsequent turns in the source conversations are not used as assistant demonstrations.

\input{tables/experiments_hyperparameters}

\begin{table}[t]
\centering\small
\caption{Task-specific training and evaluation settings. All lengths and caps are in tokenizer tokens.}
\label{tab:task-settings}
\setlength{\tabcolsep}{4pt}
\begin{tabular}{@{}>{\raggedright\arraybackslash}p{0.39\linewidth}ccc@{}}
\toprule
Setting & Math & Coding & Writing \\
\midrule
Training tasks & 3,263 & 480 & 500 \\
Epochs / final update & 2 / 180 & 16 / 208 & 16 / 208 \\
Maximum visible assistant rounds $N$ & 4 & 6 & 6 \\
Internal revision cap & 3  & 3  & 3  \\
Assistant evaluation temperature & 1 & 1 & 1 \\
Simulator evaluation temperature & 0.7 & 0.7 & 0.7 \\
Draft / prediction / revision caps & 2048 / 384 / 2048 & 1024 / 384 / 1024 & 1536 / 256 / 1536 \\
Output token limit, Prompting Base \& Naive GRPO & 9728 & 5632 & 7168 \\
Per-turn maximum user tokens $B_u$ & 332 & 2000 & 2000 \\
Per-turn maximum assistant visible tokens $B_a$ & 1024 & 600 & 900 \\
\bottomrule
\end{tabular}
\end{table}

\subsection{Interaction-Cost Counting}
\label{app:interaction-cost}

For the cost in Eq.~\eqref{eq:effort}, $n$ counts user messages, $\ell_u$ counts the
tokens of all visible user messages, including the opening request, and $\ell_a$ counts the
tokens of all visible assistant messages. Lengths are
measured in tokenizer tokens in every domain and normalized by the maxima in
Table~\ref{tab:task-settings}.

The user-effort measures in Table~\ref{tab:user-effort} use only conversations that continue past
the first assistant turn, that is, those with at least one user reply after the opening request.
For each such conversation, \emph{replies} counts the user's messages after the opening request,
and \emph{tokens} sums their lengths in Qwen3-8B tokens; when the simulator wraps its reply in a
JSON envelope, only the reply text is counted. Both are then averaged over these conversations.
A conversation has between one and $N-1$ replies, so at most three on math
(Table~\ref{tab:task-settings}). For example, \methodname{}'s 2.24 replies and 89.9 tokens on math
correspond to about 3.24 assistant turns and two short replies per conversation. Standard errors
are clustered by task (Eq.~\eqref{eq:clustered-sem}).

\subsection{Decision-Entropy Ablations}
\label{app:entropy-results}
We add $-\lambda_H H_{\mathrm{decision}}$ to the joint loss $\mathcal L$ (Eq.~\eqref{eq:joint-objective}). The entropy term is added to the GRPO gradient before the prediction-gradient cap
(Eq.~\eqref{eq:prediction-scale}) is computed, so the cap's numerator becomes
$\lVert\nabla_\theta(\mathcal L_{\mathrm{GRPO}}-\lambda_H H_{\mathrm{decision}})\rVert_2$.
For \methodname{}, the regularizer can therefore also change $\eta$, and with it the effective
weight of the prediction loss. \methodname{} task-only has $\lambda_{\mathrm{pred}}=0$ and is unaffected,
so its runs isolate the effect of the regularizer on actions alone.

\input{tables/experiments_entropy}
\input{tables/experiments_actions}
Action shares divide each action's count by the total number of recorded ASK, DRAFT, and COMMIT decisions. They pool decisions across evaluation conversations, rather than averaging conversation-level proportions, and do not measure the conditional policy entropy. Table~\ref{tab:exp:actions} shows the comparison of \methodname{} and \methodname{} task-only before and after adding entropy regularizers. The matched entropy comparisons increase ASK in all six pairs. All six have lower task scores.

\begin{samepage}
\subsection{Within-Round Self-Revision Gains on Math}
\label{app:revision-gains}
\paragraph{Metric and aggregation.}
For conversation $i$ and draft round $t$, let $d_{it}^{(0)}$ be the initial internal draft and $d_{it}^{(K_{it})}$ the final draft shown to the user. Let $s(d,z_i)$ denote strict boxed-answer correctness against the math reference. Over the set $\mathcal{R}=\{(i,t):K_{it}>0\}$ of revised rounds, we report
\begin{equation}
\Delta_{\mathrm{rev}}=\frac{100}{|\mathcal{R}|}\sum_{(i,t)\in\mathcal{R}}\left[s(d_{it}^{(K_{it})},z_i)-s(d_{it}^{(0)},z_i)\right].
\label{eq:revision-gain}
\end{equation}
\end{samepage}
This is a mean over revised rounds, not tasks or conversations. No new user message arrives between the two drafts. We use recorded drafts from completed evaluation rollouts; ASK-only rounds and rounds without a usable draft are outside the denominator.

Error bars show $\pm 1$ standard error of the mean (SEM), computed as $s_{\Delta}/\sqrt{|\mathcal{R}|}$, where $s_{\Delta}$ is the sample standard deviation of round-level score changes multiplied by 100, using denominator $|\mathcal{R}|-1$. This estimate treats revised rounds as independent and does not adjust for task or conversation clustering. Evaluation covers 365 math tasks with eight conversations each and one evaluation seed per run.

\subsection{Prediction Learning Across Checkpoints}
\label{app:prediction-learning}

\begin{figure}[t]
\centering
\includegraphics[width=0.55\linewidth]{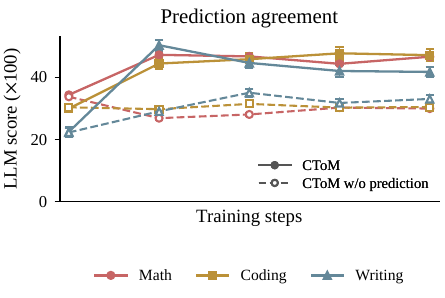}
\caption{LLM-judge agreement between predicted and observed user replies ($\times100$) for
\methodname{} (solid) and \methodname{} w/o prediction (dashed), with task-clustered SEM, at updates 0,
50, 100, 150, and a final update (180 for math, 200 for coding and writing). The first point is
the untrained model.}
\label{fig:prediction-judge}
\end{figure}

\paragraph{Runs and checkpoints.}
We evaluate the three main \methodname{} runs of Table~\ref{tab:main}, one per domain
(revision cap three, $\lambda_H=0$) at updates 0, 50, 100, 150, and the final update: 180 for
math and 208 for coding and writing. Update 0 is the untrained Qwen3-8B assistant with the same
prompt and interaction loop. We also evaluate the three \methodname{} w/o prediction runs at the same
updates; on coding, its last lexical-F1 update is 200, which
Figure~\ref{fig:prediction-learning} plots as its final point. Figure~\ref{fig:prediction-learning} plots
these five checkpoints at equal spacing in training order, and Figure~\ref{fig:prediction-judge} plots the
judge scores the same way with update 200 as the final point for coding and writing;
Table~\ref{tab:prediction-f1} also lists update 200.
Evaluation settings are fixed within each run. 

\paragraph{Prediction--reply pairs.}
At visible round $t$ the assistant shows draft $d_t$ together with its prediction $\hat u_t$ of
the user's reply, i.e.\ the last prediction made in that round's internal loop, and the simulator
then replies $u_t$. For coding and writing, the pairs of a conversation are all rounds $t$ that
have a prediction slot and an observed reply. Rounds without a prediction slot (\textsc{Ask}
and \textsc{Commit}) and a final prediction that receives no reply are excluded, and a
conversation with no such pair is excluded entirely. An empty prediction in an existing slot
scores zero. For math, the pairs are the first
$m=\min(|\hat{\mathbf u}|,\,|\mathbf u|-1)$ recorded predictions and user replies after the
opening request, matched in order; empty or missing predictions or replies score zero, and a
conversation with $m=0$ is kept with score zero.

\paragraph{Lexical F1.}
Let $W(x)$ be the set of distinct words in $x$ after
lowercasing, removing punctuation, and splitting on whitespace.
For a pair $(\hat u,u)$ with $W(\hat u)$ and $W(u)$ both non-empty,
\begin{equation}
\mathrm{F1}(\hat u,u)=\frac{2\,|W(\hat u)\cap W(u)|}{|W(\hat u)|+|W(u)|},
\label{eq:lexical-f1}
\end{equation}
which equals the harmonic mean of word-set precision $|W(\hat u)\cap W(u)|/|W(\hat u)|$ and
recall $|W(\hat u)\cap W(u)|/|W(u)|$; $\mathrm{F1}=0$ if either set is empty. Because $W$ is a
set, repeated words count once. A conversation's score is the mean F1 over its pairs, and the
reported value is the mean over eligible conversations (Eq.~\eqref{eq:clustered-sem}),
multiplied by 100.

Table~\ref{tab:prediction-f1} lists the values plotted in Figure~\ref{fig:prediction-learning}.
The two pairing conventions produce different denominators, so levels are comparable within a
run but not across domains. The math convention also lowers the untrained value: at update 0,
1{,}540 of 5{,}495 math pairs (28\%) score zero because a prediction or reply is empty or
missing, compared with 114--138 pairs (about 2\%) at later checkpoints, so part of the math
gain reflects the trained model reliably emitting a prediction.

\begin{table}[t]
\centering\small
\caption{Lexical prediction F1 (Eq.~\eqref{eq:lexical-f1}, $\times100$) of \methodname{} and
\methodname{} w/o prediction at each evaluated checkpoint, with task-clustered SEM. ``Final'' is the
checkpoint in Table~\ref{tab:main} (update 180 for math, 208 for coding and writing); a dash marks an
update that was not evaluated.}
\label{tab:prediction-f1}
\setlength{\tabcolsep}{5pt}
\begin{tabular}{@{}llcccccc@{}}
\toprule
Domain & Method & 0 & 50 & 100 & 150 & 200 & Final \\
\midrule
Math    & \methodname{} & $11.91_{\pm0.17}$ & $18.86_{\pm0.29}$ & $19.67_{\pm0.32}$ & $18.45_{\pm0.30}$ & -- & $19.70_{\pm0.31}$ \\
        & w/o prediction & $11.8_{\pm0.2}$ & $14.1_{\pm0.2}$ & $13.6_{\pm0.2}$ & $14.0_{\pm0.2}$ & -- & $14.0_{\pm0.2}$ \\
\midrule
Coding  & \methodname{} & $14.00_{\pm0.58}$ & $42.43_{\pm1.92}$ & $44.12_{\pm2.15}$ & $44.71_{\pm2.04}$ & $43.72_{\pm2.03}$ & $43.83_{\pm2.11}$ \\
        & w/o prediction & $13.8_{\pm0.5}$ & $11.4_{\pm0.5}$ & $12.8_{\pm0.6}$ & $12.4_{\pm0.5}$ & $12.8_{\pm0.5}$ & -- \\
\midrule
Writing & \methodname{} & $10.5_{\pm0.6}$ & $42.45_{\pm1.72}$ & $37.74_{\pm1.47}$ & $37.69_{\pm1.62}$ & $36.60_{\pm1.52}$ & $35.81_{\pm1.51}$ \\
        & w/o prediction & $10.4_{\pm0.6}$ & $13.8_{\pm0.7}$ & $16.3_{\pm0.7}$ & $13.8_{\pm0.6}$ & $15.6_{\pm0.8}$ & $15.8_{\pm0.7}$ \\
\bottomrule
\end{tabular}
\end{table}

\paragraph{LLM-judge agreement.}
Lexical F1 rewards shared wording rather than shared content, so we also score each pair with an
LLM judge. GPT-4o-mini at temperature 0 reads the task context, the shown draft, the predicted
reply $\hat u$, and the observed reply $u$, and rates on a 0--1 scale how well $\hat u$
anticipates the substance of $u$; the prompt is \promptref{prompt:judge}. The judge is
used only for evaluation and plays no role in training.
The judge scores exactly the pairs used for Table~\ref{tab:prediction-f1}; to confirm the
pairing, we recompute lexical F1 on the same pairs and recover the recorded values. Judge scores
are aggregated and their SEMs computed exactly as for lexical F1.
Figure~\ref{fig:prediction-judge} shows the result. In all three domains, every trained
\methodname{} checkpoint scores above the untrained model under the judge as well, so the improvement in
Figure~\ref{fig:prediction-learning} is not only a matter of copying the user's wording.
\methodname{} w/o prediction improves far less under the judge: its last checkpoint scores 29.9,
30.4, and 33.0 on math, coding, and writing, compared with 46.6, 47.1, and 41.7 for \methodname{}.

\paragraph{Standard error.}
Let task $c$ contribute $n_c$ eligible conversations whose conversation-level scores sum to
$S_c$. For $C$ tasks, the mean and its task-clustered SEM are
\begin{equation}
M=\frac{\sum_c S_c}{\sum_c n_c},\qquad
\operatorname{SEM}(M)=\frac{\sqrt{\frac{C}{C-1}\sum_c(S_c-Mn_c)^2}}{\sum_c n_c}.
\label{eq:clustered-sem}
\end{equation}
This treats tasks as independent while allowing dependence among conversations from the same
task. With equal $n_c$, it reduces to the sample standard deviation of task means divided by
$\sqrt{C}$. We use 365, 114, and 117 task clusters for math, coding, and writing; for writing,
$n_c=1$ and the SEM is the ordinary standard error over tasks. The whiskers are marginal SEMs,
not paired tests of checkpoint differences or estimates of training-seed variability.

%% file: tables/experiments_hyperparameters.tex
\begin{table}[t]
\centering\small
\caption{Shared training hyperparameters.}
\label{tab:exp:hyperparameters}
\begin{tabular}{@{}ll@{}}
\toprule
Hyperparameter & Value \\
\midrule
LoRA rank & 32 \\
GRPO Rollouts $G$ per prompt & 4 \\
Prompt batch size & 36 \\
PPO minibatch size (prompts) & 12 \\
Actor learning rate & $10^{-4}$ \\
Learning-rate schedule & Cosine \\
Warmup fraction & $0.05$ \\
Minimum learning-rate ratio & $0.1$ \\
Weight decay & $0.01$ \\
Reference-KL coefficient & $0.01$ \\
Prediction-loss coefficient (\methodname{}) & $0.1$\\
$\mathcal L_{\mathrm{pred}}$ gradient-ratio cap (\methodname{}) & $0.5$ \\
Task-reward weight & $1$ \\
Effort-reward weight & $0.01$ \\
Effort coefficients $\alpha$ / $\beta$ & $0.5$ / $0.4$ \\
Training assistant temperature & $1$ \\
Training simulator temperature & $0.7$ \\
Training simulator output cap (tokens) & 1,024 \\
Random seed & 42 \\
\bottomrule
\end{tabular}
\end{table}

%% file: tables/experiments_entropy.tex
\begin{table}[t]
\centering
\caption{Effect of the action-entropy regularizer. Each difference is the value at
$\lambda_H=10^{-3}$ minus the value at $\lambda_H=0$, using the runs in
Table~\ref{tab:exp:actions} (revision cap three in every domain). $\Delta$ASK is in
percentage points of all action decisions; $\Delta$Score is the change in task score, multiplied by 100.}
\label{tab:exp:entropy}
\begin{tabular}{llrr}
\toprule
Task & Method & $\Delta$ASK & $\Delta$Score \\
\midrule
Math    & \methodname{}  & $+22.14$ & $-7.05$ \\
        & \methodname{} task-only & $+36.77$ & $-4.24$ \\
\midrule
Coding  & \methodname{}  & $+17.75$ & $-2.12$ \\
        & \methodname{} task-only & $+15.71$ & $-4.46$ \\
\midrule
Writing & \methodname{}  & $+13.39$ & $-0.29$  \\
        & \methodname{} task-only & $+8.13$ & $-0.33$ \\
\bottomrule
\end{tabular}
\end{table}

%% file: tables/experiments_actions.tex
\begin{table}[t]
\centering\small
\caption{Observed actions and task score of each arm. ASK, DRAFT, and COMMIT are percentages of all recorded action decisions; scores are multiplied by 100, as in Table~\ref{tab:main}, with $\pm 1$ standard error of the mean. $\lambda_H$ is the action-entropy coefficient.}
\label{tab:exp:actions}
\setlength{\tabcolsep}{4pt}
\begin{tabular}{@{}llrrrrrr@{}}
\toprule
Task & Method & Inner & $\lambda_H$ & ASK & DRAFT & COMMIT & Score \\
\midrule
Math & \methodname{} & 3 & 0 & 0.00 & 97.59 & 2.41 & 79.62\pmsem{1.65} \\
 & \methodname{} & 3 & $10^{-3}$ & 22.14 & 68.29 & 9.58 & 72.57\pmsem{1.85} \\
 & \methodname{} task-only & 3 & 0 & 0.00 & 97.36 & 2.64 & 77.19\pmsem{1.75} \\
 & \methodname{} task-only & 3 & $10^{-3}$ & 36.77 & 55.44 & 7.79 & 72.95\pmsem{1.85} \\
\midrule
Coding & \methodname{} & 3 & 0 & 2.89 & 70.19 & 26.92 & 51.05\pmsem{3.08} \\
 & \methodname{} & 3 & $10^{-3}$ & 20.64 & 54.39 & 24.97 & 48.93\pmsem{3.07} \\
 & \methodname{} task-only & 3 & 0 & 27.03 & 51.65 & 21.32 & 48.11\pmsem{3.15} \\
 & \methodname{} task-only & 3 & $10^{-3}$ & 42.74 & 44.45 & 12.81 & 43.65\pmsem{3.06} \\
\midrule
Writing & \methodname{} & 3 & 0 & 32.73 & 53.19 & 14.08 & 19.08\pmsem{0.51} \\
 & \methodname{} & 3 & $10^{-3}$ & 46.12 & 32.95 & 20.93 & 18.79\pmsem{0.36} \\
 & \methodname{} task-only & 3 & 0 & 38.55 & 57.31 & 4.14 & 18.07\pmsem{0.35} \\
 & \methodname{} task-only & 3 & $10^{-3}$ & 46.68 & 50.58 & 2.75 & 17.74\pmsem{0.51} \\
\bottomrule
\end{tabular}
\end{table}